%% file: main.tex
\documentclass{article}

\usepackage{microtype}
\usepackage{graphicx}
\usepackage{subcaption}
\usepackage{booktabs} 
\usepackage{multirow} 
\usepackage{xspace}
\usepackage{xcolor}      
\usepackage{hyperref}

\usepackage{algorithm}
\usepackage{algorithmic}

\usepackage[accepted]{icml2026}
\usepackage[title]{appendix}

\usepackage{amsmath}
\usepackage{amssymb}
\usepackage{mathtools}
\usepackage{amsthm}

\usepackage[capitalize,noabbrev]{cleveref}

\theoremstyle{plain}

\theoremstyle{definition}

\theoremstyle{remark}

\usepackage[disable,textsize=tiny]{todonotes}
\usepackage{chngcntr}
\icmltitlerunning{Distilling Geometry Priors for 3D-Consistent Video Generation}

\newcommand{\methodname}{VideoGPA\xspace}

\newcommand{\videogpaurl}{%
  \href{https://hongyang-du.github.io/VideoGPA-Website/}{%
  https://hongyang-du.github.io/VideoGPA-Website
  }%
}
\begin{document}

\twocolumn[
  \icmltitle{
\methodname: Distilling Geometry Priors for 3D-Consistent Video Generation
  }



  \icmlsetsymbol{equal}{*}
  \icmlsetsymbol{equal_ad}{$\dagger$}
  
  \begin{icmlauthorlist}
    \icmlauthor{Hongyang Du}{USC,Brown,equal}
    \icmlauthor{Junjie Ye}{USC,equal}
    \icmlauthor{Xiaoyan Cong}{Brown,equal}
    \icmlauthor{Runhao Li}{USC}
    \icmlauthor{Jingcheng Ni}{Brown} \\
    \icmlauthor{Aman Agarwal}{Brown}
    \icmlauthor{Zeqi Zhou}{Brown}
    \icmlauthor{Zekun Li}{Brown}
    \icmlauthor{Randall Balestriero}{Brown,equal_ad}
    \icmlauthor{Yue Wang}{USC,equal_ad} \\
    \videogpaurl

  \end{icmlauthorlist}
  
  \icmlaffiliation{USC}{Physical Superintelligence Lab, University of Southern California, Los Angeles, CA, US}
  \icmlaffiliation{Brown}{Department of Computer Science, Brown University, Providence, RI, US}

  \icmlcorrespondingauthor{Hongyang Du}{hongyang\_du@brown.edu}
  \icmlcorrespondingauthor{Junjie Ye}{yejunjie@usc.edu}

  \icmlkeywords{Machine Learning, ICML}

  \vskip 0.3in
]



\printAffiliationsAndNotice{\icmlEqualContribution \icmlEqualAdvising}

\input{sections/00-abstract}
\label{abstract}

\vspace{-13pt}

\begin{figure}[!t]
      \centerline{\includegraphics[width=\linewidth]{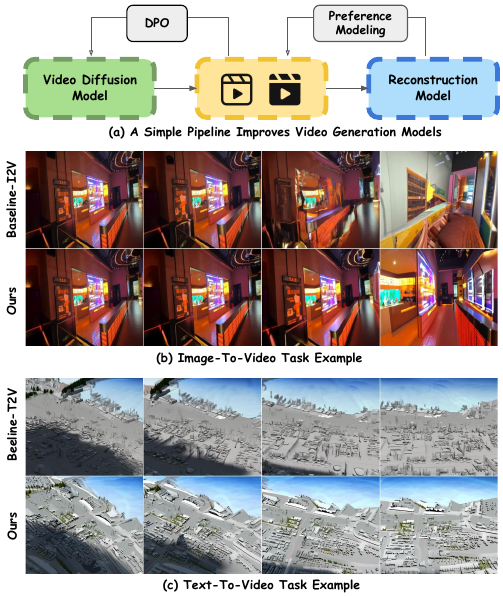}}
    \caption{\textbf{Overview of \methodname and representative results.} 
    (a) \methodname aligns a pretrained video diffusion model through proposed reconstruction-guided preference optimization.
    (b) Image-to-video examples comparing the base model~\cite{yang2025cogvideoxtexttovideodiffusionmodels} and VideoGPA, showing improved geometric stability under camera motion. 
    (c) Text-to-video examples demonstrating improved structural coherence and reduced geometric artifacts.
    }
  \label{fig:Teaser}
  \vspace{-15pt}
\end{figure}

\input{sections/introduction_xiaoyan}

\input{sections/20-relatedwork}

\input{sections/30-method}

\input{sections/40-experiments}


\input{sections/50-discussion}


\input{sections/70-conclusion}


\bibliography{main}
\bibliographystyle{icml2026}


\input{sections/appendix}

\label{appendix}


\end{document}

%% file: sections/00-abstract.tex
\begin{abstract}

While recent video diffusion models (VDMs) produce visually impressive results, they fundamentally struggle to maintain 3D structural consistency, often resulting in object deformation or spatial drift. 
We hypothesize that these failures arise because standard denoising objectives lack explicit incentives for geometric coherence. 
To address this, we introduce \textbf{\methodname} (\textbf{Video} \textbf{G}eometric \textbf{P}reference \textbf{A}lignment), a data-efficient self-supervised framework that leverages a geometry foundation model to automatically derive dense preference signals that guide VDMs via Direct Preference Optimization (DPO). This approach effectively steers the generative distribution toward inherent 3D consistency without requiring human annotations. \methodname significantly enhances temporal stability, geometric plausibility, and motion coherence using minimal preference pairs, consistently outperforming state-of-the-art baselines in extensive experiments.

\end{abstract}

%% file: sections/introduction_xiaoyan.tex
\section{Introduction}

The rapid evolution of Video Diffusion Models (VDMs) has revolutionized content creation, achieving remarkable generalizability and visual fidelity~\cite{hong2022cogvideo,yang2025cogvideoxtexttovideodiffusionmodels,kong2025hunyuanvideosystematicframeworklarge, wan2025wanopenadvancedlargescale,nvidia2025cosmosworldfoundationmodel, seedance2025seedance15pronative}. 
Beyond artistic generation, the community is actively exploring the potential of VDMs as data engines for downstream tasks such as Embodied AI~\cite{bruce2024genie,bruce2024geniegenerativeinteractiveenvironments,yang2024learninginteractiverealworldsimulators, feng2025vidar,li2025novaflowzeroshotmanipulationactionable,  ye2025anchordreamrepurposingvideodiffusion, deng2026rethinking,kim2026cosmospolicyfinetuningvideo}, Novel View Synthesis (NVS)~\cite{liu2023zero1to3zeroshotimage3d,  yu2024viewcraftertamingvideodiffusion, kwak2024vivid, voleti2024sv3dnovelmultiviewsynthesis,you2025nvssolvervideodiffusionmodel}, and physics simulation~\cite{sun2024dimensionxcreate3d4d,cai2024generativevideomodelshelp,qin2024worldsimbenchvideogenerationmodels,liu2024physics3dlearningphysicalproperties,chen2025first}. Across these applications, faithful 3D understanding plays a fundamental role.
This raises a fundamental question: \textit{Have these models truly learned the 3D laws of the world?} We posit that a 3D consistent VDM, which inherently respects geometric constraints, is critical to model the 3D world.

However, despite being pre-trained on billion-scale datasets, current VDMs still exhibit deficiencies in 3D consistency. 
As illustrated in Fig.~\ref{fig:Teaser}, pre-trained VDMs fail to maintain structural consistency and temporal stability. 
Common artifacts include object deformation, spatial drifting, and geometry collapse over time.

We attribute this paradox, where models have seen scalable 3D-consistent data but exhibit inconsistent behaviors, to the nature of denoising objective. 
Standard training incentivizes pixel-level statistical matching but lacks geometric regularization. 
Consequently, the model learns to hallucinate plausible textures without effectively injecting 3D consistency into latent space.

Recent advances in Geometry Foundation Models (GFMs) demonstrate strong geometric priors through their ability to infer dense 3D structure and camera motion from 2D observations~\cite{dust3r,wang2025vggt}. We leverage these priors by distilling reconstruction-based 3D knowledge into video diffusion models, aligning generation toward physically consistent geometry without retraining from scratch or relying on human annotations.


In this work, we introduce \textbf{\methodname} (\textbf{Video} \textbf{G}eometric \textbf{P}reference \textbf{A}lignment), a data-efficient, self-supervised framework that equips pre-trained VDMs with 3D consistency. 
The key to our approach is a novel 3D consistency metric derived from the principle of re-projection consistency. 
Specifically, given a generated video, we utilize a GFM to render a 3D consistent video as a reference.
The discrepancy between the input video and the rendered reference serves as a robust proxy for 3D consistency: if a video is geometrically valid, the GFM-derived 3D structure should accurately reconstruct the original input.
By leveraging a geometry foundation model as a reward model backbone, we construct geometric preference pairs, distinguishing between samples with high and low structural integrity. 
These pairs guide the VDM via Direct Preference Optimization (DPO)~\cite{rafailov2024directpreferenceoptimizationlanguage}, effectively steering the generative distribution toward the 3D-consistent manifold. 
Remarkably, we show that with only $\sim 2,500$ preference pairs and minimal post-training (LoRA fine-tuning~\cite{hu2021loralowrankadaptationlarge} on $\sim$1\% of model parameters), \methodname substantially improves geometric coherence and temporal stability, while preserving the base model’s visual quality and motion realism. Across both image-to-video and text-to-video settings, \methodname consistently improves over prior art on multiple geometric consistency and perceptual metrics.


%% file: sections/20-relatedwork.tex
\section{Related Works}
\subsection{Video Generation Models}
Recent video generation has achieved high visual fidelity by scaling Diffusion Transformer (DiT) architectures~\cite{yang2025cogvideoxtexttovideodiffusionmodels, kong2025hunyuanvideosystematicframeworklarge, wan2025wanopenadvancedlargescale, seedance2025seedance15pronative, nvidia2025cosmosworldfoundationmodel}. Within this paradigm, \citet{wan2025wanopenadvancedlargescale} optimize for high-ratio temporal compression via a 3D causal VAE, while \citet{kong2025hunyuanvideosystematicframeworklarge} introduce a hybrid dual-stream design to refine multimodal fusion. 
CogVideoX~\citep{yang2025cogvideoxtexttovideodiffusionmodels} serves as a representative of this class, utilizing expert transformer blocks to model spatiotemporal patches. Despite these achievements, these models are fundamentally optimized for pixel-level denoising rather than geometric structural coherence. Consequently, they struggle with 3D geometric consistency, often exhibiting object deformation or spatial drift under global camera maneuvers. 


\subsection{Video Diffusion Alignment}
To address these structural bottlenecks, recent research adapts post-training alignment frameworks to refine the output manifold of diffusion models. These approaches generally follow the paradigms of supervised fine-tuning (SFT) and reinforcement learning (RL). SFT-based methods such as Force Prompting~\cite{gillman2025forcepromptingvideogeneration} improve consistency by training on high-quality curated data, yet they often suffer from limited generalization. Alternatively, RL-based approaches like DDPO~\cite{black2024trainingdiffusionmodelsreinforcement}, Flow-GRPO~\cite{liu2025flowgrpotrainingflowmatching}, and DanceGRPO~\cite{xue2025dancegrpounleashinggrpovisual} frame denoising as a multistep decision process to optimize for aesthetic or motion rewards. Diffusion-DPO~\cite{wallace2023diffusionmodelalignmentusing} further simplifies this by providing a stable offline objective for preference learning without the complexity of iterative sampling. While \citet{kupyn2025epipolar} explore geometric alignment via epipolar constraints and \citet{liu2025improvingvideogenerationhuman} utilize human feedback, we introduce a self-supervised paradigm that leverages feed-forward 3D reconstruction as a dense automated geometric signal.

\subsection{Geometric Foundation Models}
Recent advances in geometric foundation models offer potential for recovering dense geometric structure from sparse views without iterative optimization~\cite{Du_2026_CVPR}. This paradigm, established by DUSt3R~\cite{dust3r}, utilizes transformer architectures to regress pointmaps directly from pixels, while subsequent models like MASt3R~\cite{mast3r} further refine local matching. More recently, \citet{wang2025vggt} and \citet{wang2025pi3permutationequivariantvisualgeometry} scale this approach to multi-view sequences by simultaneously predicting globally consistent camera poses and pointmaps. This capability provides a stable, dense, and differentiable 3D reference, bridging the gap between 2D appearance and 3D space. Such models offer huge potential as automated geometric supervisors for generative tasks. By distilling these structural priors into the diffusion process, we explore how reconstruction can help rectify the motion manifold of a generator to ensure geometric plausibility while maintaining the simplicity of the standard generative pipeline.

%% file: sections/30-method.tex
\section{Methodology}
\label{method}
\begin{figure}[t]
  \centerline{\includegraphics[width=\linewidth]{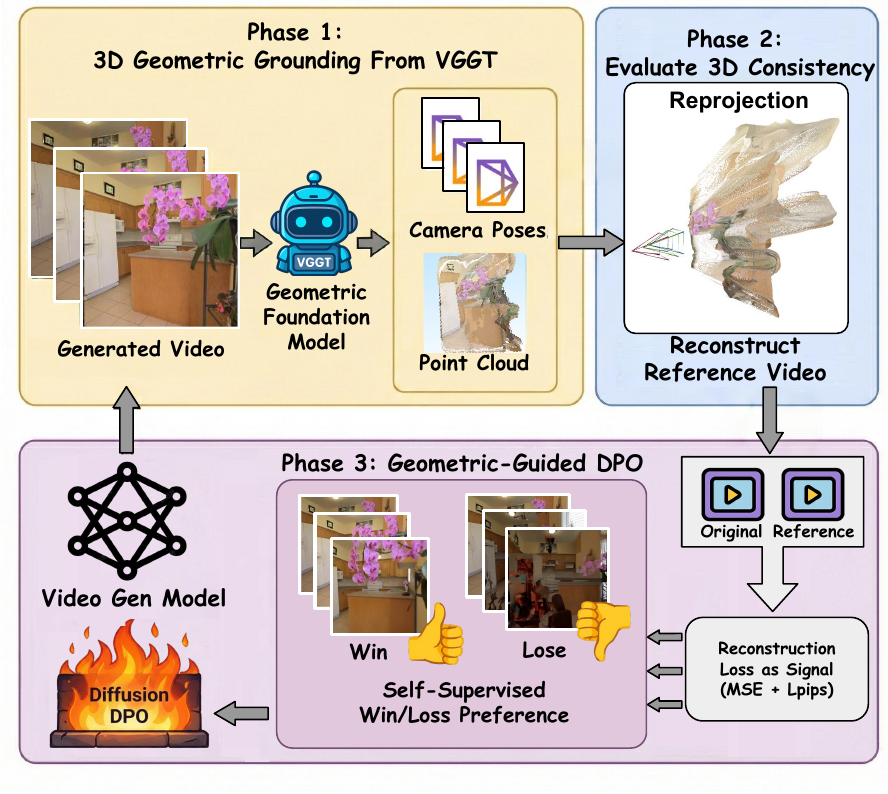}}
  \caption{\textbf{Pipeline of \methodname.} A geometric foundation model probes generated videos to assess scene-level 3D consistency, which is used to form self-supervised preference pairs for post-training alignment via DPO.}
  \label{fig:pipeline}
  \vspace{-12pt}
\end{figure}

VideoGPA introduces a review-and-correct framework that aligns video diffusion models with 3D geometric laws. As shown in Fig.~\ref{fig:pipeline}, the process begins by using a geometry foundation model to extract the 3D structure and camera motion from generated videos. We then calculate a 3D consistency score by measuring the error in reconstructing the original frames from this 3D structure. Finally, with preference pairs constructed using these scores, we apply DPO to teach the model to favor geometrically consistent outputs. This lightweight approach allows the model to maintain rigid structures with minimal additional training.
\subsection{Preliminaries}

\paragraph{Direct Preference Optimization}
The DPO~\cite{rafailov2024directpreferenceoptimizationlanguage} framework is derived from the Bradley-Terry preference model. For a policy $\pi$, the probability that a preferred completion $x^w$ is chosen over a dispreferred completion $x^l$ given a context $c$ is:
\begin{equation}
    P(x^w \succ x^l | c) = \sigma \left( r(x^w, c) - r(x^l, c) \right),
\end{equation}
where $r(x, c)$ is the reward function and $\sigma$ is the sigmoid function. DPO reparameterizes the reward using the log-likelihood ratio between policy $\pi_\theta$ and reference policy $\pi_{\text{ref}}$:
\begin{equation}
    r(x, c) = \beta \log \frac{\pi_\theta(x | c)}{\pi_{\text{ref}}(x | c)} + \beta \log Z(c).
\end{equation}
Substituting this into the Bradley-Terry model yields the DPO objective:
\begin{equation}
\begin{split}
    \mathcal{L}_{\text{DPO}} = -\mathbb{E}_{(c, x^w, x^l)} \Big[ \log \sigma \Big( &\beta \log \frac{\pi_\theta(x^w | c)}{\pi_{\text{ref}}(x^w | c)} \\
    &- \beta \log \frac{\pi_\theta(x^l | c)}{\pi_{\text{ref}}(x^l | c)} \Big) \Big].
\end{split}
\end{equation}
\paragraph{From Policy to Diffusion Log-likelihood}
In diffusion models, the log-probability $\log p(x)$ is typically approximated via the Evidence Lower Bound (ELBO). For a denoising model $\epsilon_\theta$, the log-probability ratio can be expressed as the difference in the score-matching loss~\cite{wallace2023diffusionmodelalignmentusing}. Specifically, for a given timestep $t$:
\begin{equation}
\begin{split}
    \log \frac{\pi_\theta(x | c)}{\pi_{\text{ref}}(x | c)} \propto - \mathbb{E}_{t, \epsilon} \bigg[ &\|\epsilon - \epsilon_\theta(x_t, t, c)\|^2 \\
    &- \|\epsilon - \epsilon_{\text{ref}}(x_t, t, c)\|^2 \bigg],
\end{split}
\end{equation}
where $x_t = \sqrt{\bar{\alpha}_t}x_0 + \sqrt{1-\bar{\alpha}_t}\epsilon$ represents the noisy latent at timestep $t$.

\subsection{DPO for $v$-Prediction Video Diffusion}

Recent advancements in video generation have been significantly driven by diffusion transformers (DiTs) employing the $v$-prediction parameterization~\cite{salimans2022progressivedistillationfastsampling, ho2022video,lipman2023flowmatchinggenerativemodeling}. Trained on large-scale datasets, these models demonstrate remarkable potential in synthesizing high-fidelity dynamic content with stable training convergence. Building upon this foundation, we adapt the Diffusion-DPO~\cite{wallace2023diffusionmodelalignmentusing} framework to this parameterization to explicitly steer the model toward geometrically consistent manifolds.

For a general diffusion model trained with $v$-prediction, the target velocity $v_t$ at timestep $t$ is formally defined as:
\begin{equation}
    v_t \equiv \dot{x}_t = \alpha_t \epsilon - \sigma_t x_0,
\end{equation}
where $\alpha_t$ and $\sigma_t$ are the noise schedule coefficients. The network $v_\theta$ is trained to minimize the mean squared error (MSE) relative to the velocity target $v_t$. 

By substituting the velocity error into the DPO framework, we define the energy term $\mathcal{E}$ for a sample $x$ as:
\begin{equation}
    \mathcal{E}(\theta, x, t) = \|v_t - v_\theta(x_t, t, c)\|^2.
\end{equation}
The log-probability ratio in the velocity space thus becomes:
\begin{equation}
    \log \frac{\pi_\theta(x | c)}{\pi_{\text{ref}}(x | c)} \propto \mathbb{E}_{t, \epsilon} \left[ \mathcal{E}(\text{ref}, x, t) - \mathcal{E}(\theta, x, t) \right]
\end{equation}
The final DPO loss for our $v$-prediction video model is formulated as the negative log-likelihood of the reward margin between the winning and losing samples:
\begin{equation}
\begin{split}
    \mathcal{L}_{\text{DPO}} = -\mathbb{E} \bigg[ \log &\sigma \Big( \beta \Big([\mathcal{E}(\text{ref}, x^w, t) - \mathcal{E}(\theta, x^w, t)] \\
    &- [\mathcal{E}(\text{ref}, x^l, t) - \mathcal{E}(\theta, x^l, t)] \Big) \Big) \bigg].
\end{split}
\end{equation}
During training, we sample shared noise $\epsilon$ and timestep $t$ for each preference pair $(x^w, x^l)$ to ensure a consistent optimization baseline. The model $v_\theta$ (parameterized via LoRA~\cite{hu2021loralowrankadaptationlarge}) is updated to minimize the velocity prediction error for $x^w$ relative to $x^l$, effectively steering the latent video manifold toward higher geometric consistency.

\subsection{Preference Modeling}
\label{sec:consistency score}
DPO requires preference pairs to guide post-training alignment. In our setting, the objective favors video samples that exhibit stronger 3D geometric consistency. To this end, we derive a self-supervised geometric preference signal by analyzing generated videos with a feed-forward geometric foundation model~\cite{wang2025vggt}. The resulting 3D consistency score enables automatic construction of preference pairs without human annotations or explicit structural priors. We next describe how this score is computed from reconstructed camera motion and scene geometry.

For each generated video, we uniformly sample $T$ frames to obtain an image sequence $\mathcal{I}=\{I_t\}_{t=1}^{T}$.
Given this sequence, the geometric model $\Phi$ predicts a depth map $D_t$ and camera pose $(R_t, t_t)$ for each frame $I_t$,
\begin{equation}
(D_t, R_t, t_t) = \Phi_{\theta}(I_t),
\qquad
R_t \in SO(3),\; t_t \in \mathbb{R}^3,
\label{eq:vggt_final}
\end{equation}
along with camera intrinsics $K\!\in\!\mathbb{R}^{3\times3}$.
Using the camera-to-world transform $E_t=[R_t|t_t]\!\in\!SE(3)$,
each pixel $\tilde{\mathbf{u}}=[u,v,1]^{\top}$ with depth $D_t(u,v)>0$ can be projected to the world coordinate frame with:
\begin{equation}
\begin{aligned}
\mathbf{x}_t^{\mathrm{cam}}(u,v) &= D_t(u,v)\,K^{-1}\tilde{\mathbf{u}},\\
\mathbf{X}_t(u,v) &= R_t\,\mathbf{x}_t^{\mathrm{cam}}(u,v)+t_t,
\end{aligned}
\label{eq:backproj_final}
\end{equation}
formulating a colored point cloud
$\mathcal{P}=\{(\mathbf{X}_i,\mathbf{c}_i)\}_{i=1}^{N}$,
where $\mathbf{c}_i$ is RGB color. By default, we set $T = 10$.

\paragraph{3D Consistency Score}
We measure 3D geometric consistency by how well the recovered structure explains the original frames under reprojection. Geometrically coherent videos admit a consistent 3D explanation across viewpoints, while inconsistencies lead to elevated reprojection error.

Formally, each 3D point $\mathbf{X}\!\in\!\mathcal{P}$ is reprojected into frame $k$
using inverse camera pose $E_{t,\mathrm{w2c}}=[R_t^{\top}|-R_t^{\top}t_t]$, yielding 
\begin{equation}
\mathbf{x}_t^{\mathrm{cam}}=R_t^{\top}(\mathbf{X}-t_t),
\quad
(u_t,v_t)=\pi(K\,\mathbf{x}_t^{\mathrm{cam}}),
\label{eq:reproj_final}
\end{equation}
where $\pi$ denotes perspective division.
Using vectorized rendering with a painter’s algorithm, we obtain a reprojected image $\hat{I}_t$ for each frame. 

We quantify geometric consistency between reprojected image $\{\hat{I}_t\}_{t=1}^T$ and original frame $\{I_t\}_{t=1}^T$
using a standard reconstruction loss~\cite{yao2018mvsnetdepthinferenceunstructured}:
\begin{equation}
\label{eq:3D_LOSS_final}
E_{\rm Recon} = \frac{1}{T}\sum_{t=1}^{T}
\Big(
\mathrm{MSE}\big(\hat{I}_t,\, I_t\big)
+
\lambda \mathrm{LPIPS}\big(\hat{I}_t,\, I_t\big)
\Big).
\end{equation}
Lower reconstruction error indicates stronger cross-view geometric consistency, while higher error reflects violations of 3D coherence. We use this reprojection-based error as a dense, self-supervised signal to construct preference pairs.

\subsection{Post-Training Alignment}\label{subsec:preference_data}
With the proposed preference modeling strategy, we curate a specialized training dataset for post-training alignment by exposing geometric differences among candidate generations. Specifically, for each conditioning input, we sample multiple videos from a pretrained video diffusion model using different random seeds. These samples share the same semantic content but may differ in geometric consistency, allowing us to isolate geometry as the primary factor for preference learning and construct preference pairs based on the 3D consistency score (Sec.~\ref{sec:consistency score}). We consider both image-to-video and text-to-video settings to ensure the learned alignment generalizes across conditioning modalities.

\paragraph{Image-to-Video (I2V)}
For I2V data, we use the initial frames from a subset of DL3DV-10K \cite{ling2023dl3dv10klargescalescenedataset} as visual prompts. To encourage the model to generate samples with diverse camera trajectories where geometric inconsistencies are more likely to emerge, we design structured motion prompts composed of 2–3 randomly sampled camera motion primitives from a predefined vocabulary (\textit{e.g.}, \emph{pull back away from the scene}, \emph{roll gently to one side}, \emph{orbit around the scene}). The full list of motion primitives is provided in Appendix~\ref{sec:appendix_data}. 
This scripted prompting strategy systematically varies camera motion while keeping scene content fixed, making geometric consistency the dominant distinguishing factor among candidate samples and reducing confounding effects from semantic variation.

\paragraph{Text-to-Video (T2V)}
For T2V data, we use video captions generated by CogVLM2-Video~\cite{hong2024cogvlm2visuallanguagemodels} as textual prompts. Compared to the I2V setting, this setup naturally introduces higher semantic diversity and more open-ended scene dynamics, enabling us to evaluate whether the proposed alignment strategy remains effective under unconstrained language inputs and complex scene compositions.

For each prompt, we rank candidate samples according to their 3D consistency scores and form preference pairs by selecting samples with a sufficient margin in geometric consistency. To ensure a stable and informative training signal, we prune samples that are static, exhibit poor overall visual quality, or show negligible score differences, following the filtering strategy detailed in Appendix~\ref{sec:appendix_data}.

\input{tables/quantitative_evaluation}

During evaluation, we adopt natural, descriptive narrations generated by CogVLM2-Video~\cite{hong2024cogvlm2visuallanguagemodels} as prompts for both I2V and T2V tasks. Although this differs from the structured motion prompts used during I2V training, we observe no evidence of overfitting to the scripted prompt format. Instead, models trained with our alignment strategy retain their ability to handle detailed natural language descriptions while exhibiting substantially improved 3D geometric consistency.

%% file: tables/quantitative_evaluation.tex
\begin{table*}[!t]
\small
\centering
\caption{\textbf{Quantitative evaluation} on image-to-video (I2V) and text-to-video (T2V) generation. We report 3D reconstruction error, 3D geometric consistency metrics, and human-aligned VideoReward scores. 
Overall, \methodname consistently achieves the strongest geometric consistency while maintaining or improving perceptual quality across various base models.
}
\label{tab:final_3d_comparison}
\begin{tabular}{l|ccc|ccc|cccc}
\toprule
\multirow{2}{*}{\textbf{Method}} & \multicolumn{3}{c|}{\textbf{3D Reconstruction Error}} & \multicolumn{3}{c|}{\textbf{3D Consistency}} & \multicolumn{4}{c}{\textbf{VideoReward (Win Rate \%)}} \\
\cmidrule(lr){2-4} \cmidrule(lr){5-7} \cmidrule(lr){8-11}
 & \textbf{PSNR} $\uparrow$ & \textbf{SSIM} $\uparrow$ & \textbf{LPIPS} $\downarrow$ & \textbf{MVCS} $\uparrow$ & \textbf{3DCS} $\downarrow$ & \textbf{Epipolar} $\downarrow$ & \textbf{VQ} & \textbf{MQ} & \textbf{TA} & \textbf{OVL} \\
\midrule
 \multicolumn{11}{c}{Image-to-Video (I2V) \textit{Base Model: CogVideoX-I2V-5B}} \\
\midrule
Baseline-I2V & \textbf{22.85} & \textbf{0.786} & 0.476 & 0.945 & 0.485 & 0.585 & - & - & - & - \\
SFT & 21.58 & 0.749 & 0.513 & 0.947 & 0.524 & 0.640 & 44.67 & 33.00 & 52.67 & 35.00 \\
Epipolar-DPO & 21.38 & 0.773 & 0.475 & 0.944 & 0.487 & 0.545 & 67.33 & 51.33 & 56.67 & 66.00\\
\textbf{VideoGPA (Ours)} & 21.24 & 0.779 & \textbf{0.473} & \textbf{0.950} & \textbf{0.483} & \textbf{0.539} & \textbf{74.00} & \textbf{56.00} & \textbf{57.67} & \textbf{76.00} \\
\midrule
\multicolumn{11}{c}{Text-to-Video (T2V) \textit{Base Model: CogVideoX-5B}} \\
\midrule
Baseline-T2V & 21.47 & 0.784 & 0.435 & 0.944 & 0.445 & 0.584 & - & - & - & - \\
SFT & 19.99 & 0.721 & 0.496 & 0.937 & 0.510 & 0.719 & 14.67 & 23.67 & 39.33 & 15.33 \\
Epipolar-DPO & \textbf{21.58} & 0.791 & 0.434 & \textbf{0.953} & 0.443 & 0.579 & 45.00 & 53.67 & 49.00 & 48.67 \\
\textbf{VideoGPA (Ours)} & 21.24 & \textbf{0.803} & \textbf{0.411} & \textbf{0.953} & \textbf{0.422} & \textbf{0.548} & \textbf{62.67} & \textbf{67.00} & 42.67& \textbf{60.33} \\
\midrule
\multicolumn{11}{c}{Text-to-Video (T2V) \textit{Base Model: CogVideoX1.5-5B}} \\
\midrule
Baseline-T2V15 & 16.79 & 0.538 & 0.522 & 0.980 & \textbf{0.548} & 0.685 & - & - & - & - \\
GeoVideo & 15.20 & 0.458 &0.667 & 0.819 & 0.703 & 0.875 & 17.36 & 44.44 & 30.56 & 18.06 \\
\textbf{VideoGPA (Ours)} & 14.88 & 0.503 & \textbf{0.520} & \textbf{0.982} &  0.556 & \textbf{0.567} & \textbf{60.42} & \textbf{54.17} & \textbf{52.08}& \textbf{57.64} \\

\bottomrule
\end{tabular}
\raggedright \footnotesize \textit{Note: All reprojection based metrics are calculated using the Depth Anything V3 backbone to prevent circular evaluation.}
\vspace{-7pt}
\end{table*}

%% file: sections/40-experiments.tex
\section{Experiments}
\label{section:experiments}

\subsection{Experimental Setup}
\subsubsection{Training Details}
We evaluate our method using CogVideoX~\cite{yang2025cogvideoxtexttovideodiffusionmodels}, a representative large-scale video diffusion model, in both I2V and T2V settings. We fine-tune CogVideoX 5B models using LoRA, employing a rank $r=64$ and a scaling factor $\alpha=128$ within the PEFT framework (approximately 1\% of model parameters)~\cite{hu2021loralowrankadaptationlarge, xu2023parameterefficientfinetuningmethodspretrained}. Training is performed on 8$\times$A100 GPUs for 10,000 steps using AdamW optimizer, with a peak learning rate of $5 \times 10^{-6}$, a cosine decay schedule, 500 warm-up steps, and batch size 16. Unless otherwise specified, subsets $8K$, $9K$, $10K$, and $11K$ from DL3DV-10K~\cite{ling2023dl3dv10klargescalescenedataset} are used for training, and subset $1K$ is used for evaluation.

\subsubsection{Baselines}
We compare our proposed \methodname against several representative baselines that reflect different strategies for improving geometric consistency in video generation:

\textit{Base model} uses the pretrained video diffusion model~\cite{yang2025cogvideoxtexttovideodiffusionmodels} without post-training alignment as a reference for evaluating the effect of alignment.

\textit{Supervised Fine-Tuning (SFT)} fine-tunes the base video diffusion model on curated video-text pairs. This baseline represents data-driven improvement without explicit geometric preference modeling.

\textit{Epipolar-DPO}~\cite{kupyn2025epipolar} applies DPO using epipolar errors as a preference signal. This method leverages sparse pair-wise geometric constraints to rank samples.

\textit{GeoVideo}~\cite{bai2025geovideointroducinggeometricregularization} incorporates explicit geometric consistency losses during supervised fine-tuning to encourage stable scene structure in large camera motion.

For fair comparison, Epipolar-DPO and SFT are trained using the same LoRA configuration and optimization schedule as our method. Detailed GPU usage is in Appendix~\ref{sec:GPU}.

\subsubsection{Evaluation Protocol}
We evaluate all methods along three complementary dimensions to capture geometric fidelity and perceptual quality:

\paragraph{3D Reconstruction Error} We assess the pixel-level fidelity of the video-to-3D reconstruction process by reprojecting reconstructed geometry back to the image plane. Quantitative metrics include PSNR for intensity differences, SSIM for structural similarity, and LPIPS for perceptual similarity between reprojected frames and the original video.
\paragraph{3D Consistency} We measure the geometric plausibility and cross-view consistency of generated videos using multiple geometry-based metrics. Multi-View Consistency Score (MVCS)~\cite{bai2025geovideointroducinggeometricregularization} evaluates cross-frame projection accuracy, our 3D Consistency Score (3DCS) measures global scene integration as defined in Eq.~\ref{eq:3D_LOSS_final}, and the Epipolar Sampson Error~\cite{kupyn2025epipolar} quantifies point-to-epipolar-line distances, reflecting camera-to-scene geometric precision.
\paragraph{Human-Aligned Video Quality} To evaluate alignment with human preferences, we use VideoReward~\cite{liu2025improvingvideogenerationhuman} to compute win rates against baselines. This includes Visual Quality (VQ), Motion Quality (MQ), Text Alignment (TA), and Overall (OVL) scores.
Please note that static videos are filtered out during both training and evaluation to ensure reported metrics accurately reflect performance on the motion manifold rather than stationary content.

\subsection{Quantitative Evaluation}
\paragraph{Image-to-Video (I2V)}We first evaluate I2V generation, where maintaining geometric consistency relative to the input frame is particularly challenging. As shown in Table~\ref{tab:final_3d_comparison}, \methodname consistently improves both geometric fidelity and perceptual quality over all baselines. For the \textbf{CogVideoX-I2V-5B} baseline, \methodname achieves the best performance across all 3D consistency metrics, improving MVCS from $0.945$ to $0.950$, reducing 3DCS from $0.485$ to $0.483$, and decreasing Epipolar error from $0.585$ to $0.539$. These geometric gains translate to significantly stronger human-aligned performance, reaching an overall VideoReward (OVL) win rate of $76.0\%$, substantially outperforming Epipolar-DPO ($66.0\%$) and SFT ($35.0\%$).

\paragraph{Text-to-Video (T2V)}Without a reference image, T2V introduces greater semantic diversity. As shown in Table~\ref{tab:final_3d_comparison} (middle), \methodname again delivers consistent improvements over the \textbf{CogVideoX-5B} base model and competing baselines. In terms of 3D reconstruction error, \methodname achieves the best SSIM (0.803) and LPIPS (0.411). It also outperforms Epipolar-DPO and SFT across all 3D consistency metrics, achieving the highest MVCS (0.953) and the lowest 3DCS (0.422) and epipolar error (0.548). Importantly, these geometric gains do not come at the cost of perceptual quality. \methodname substantially improves human-aligned metrics, achieving an overall VideoReward win rate of 60.33\%, compared to 48.67\% for Epipolar-DPO and 15.33\% for SFT. These results indicate that \methodname generalizes beyond image-conditioned generation and remains effective under unconstrained text prompts.

We further compare \methodname with GeoVideo \cite{bai2025geovideointroducinggeometricregularization} in the T2V setting. Since GeoVideo is based on \textbf{CogVideoX1.5-5B}, we post-train the same base model with \methodname. Notably, \methodname is trained for only 1,500 steps, in contrast to GeoVideo, which is trained on $\sim$10,000 DL3DV-10K videos with depth supervision. 
As shown in Table~\ref{tab:final_3d_comparison}, GeoVideo improves reconstruction accuracy for large camera motions, but exhibits degraded perceptual quality, reflected by low VideoReward scores (OVL: 18.06\%). In contrast, with this lightweight post-training, \methodname already maintains the base model's generative quality while achieving stronger overall geometric consistency (Epipolar: 0.567 v.s. 0.875; MVCS: 0.982 v.s. 0.819) and substantially higher human-aligned performance (OVL: 57.64\%). These results suggest that geometry-guided preference alignment provides a more balanced improvement than explicit geometric supervision.

\paragraph{Additional Quantitative Evaluation}
We further verify that \methodname does not sacrifice scene complexity or motion dynamics, as detailed in Appendix~\ref{sec:scene_complexity}. Additional out-of-distribution (OOD) evaluation results on the WebVid~\cite{bain2021frozen} and Panda-70M~\cite{chen2024panda} datasets are presented in Appendix~\ref{sec:ood_evaluation}. We further evaluate Wan2.2~\cite{wan2025wanopenadvancedlargescale} to examine \methodname generalizability, reported in 
Appendix~\ref{sec:wan}.  Finally, comprehensive visual comparisons and qualitative examples are provided in Appendix~\ref{sec:qualititive_results}.

\begin{figure}[t]
  \centering
  \includegraphics[width=\linewidth]{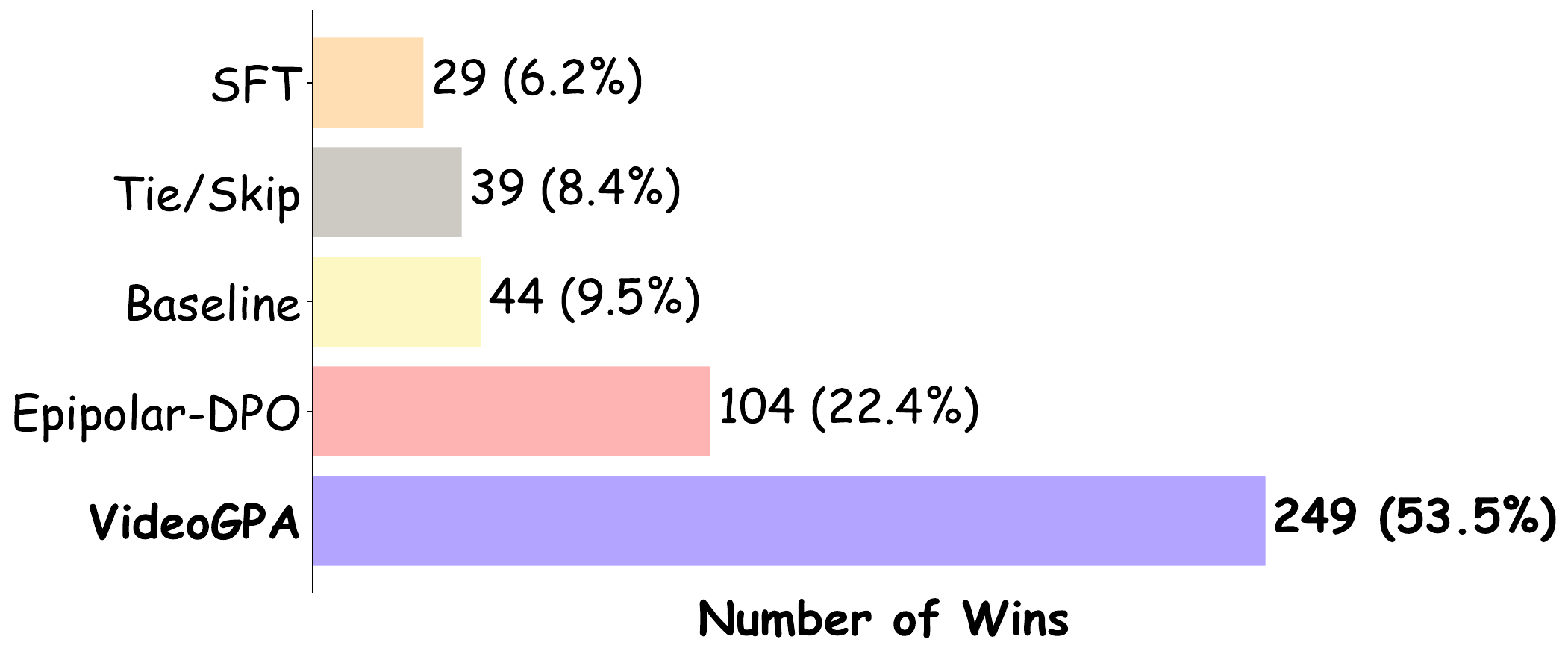}
  \caption{\textbf{Human preference study} on I2V generation. \methodname is most frequently preferred, indicating improved perceptual quality and 3D consistency.
  }
  \label{fig:user_study}
  \vspace{- 5pt}
\end{figure}

\begin{figure*}[t]
  \centerline{\includegraphics[width=\textwidth]{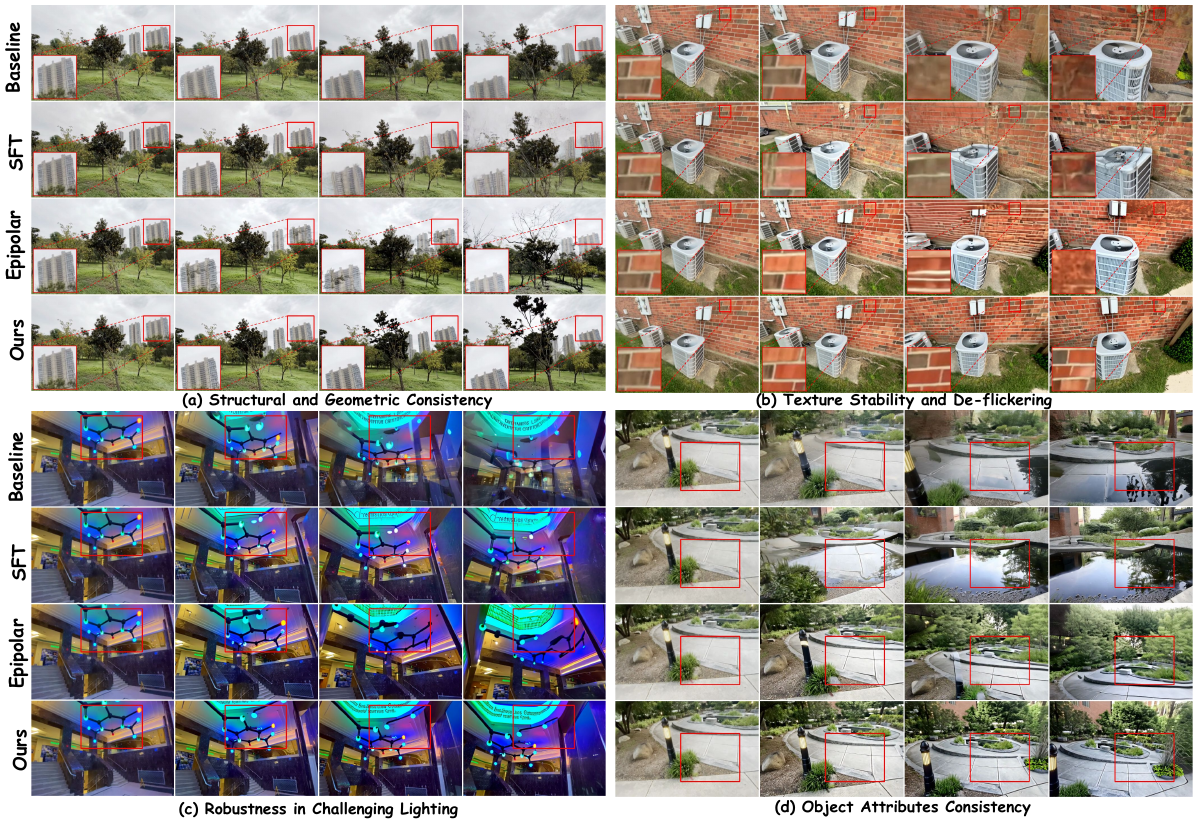}}
 \caption{
 \textbf{Qualitative comparison} on I2V generation. We compare \methodname with the base model, SFT, and Epipolar-DPO. Highlighted regions illustrate improvements in (a) structural and geometric consistency, (b) texture stability and de-flickering, (c) robustness under challenging lighting, and (d) object attribute consistency.
}
  \label{fig:i2v teaser}
  \vspace{-8pt}
\end{figure*}

\subsection{Human Preference Study}

While the quantitative results demonstrate consistent improvements across metrics, we further evaluate whether these gains are perceptible to humans through a blind user preference study. The study is performed in the I2V setting. A total of 25 participants are each assigned 20 randomly sampled video groups, where each group contains four videos generated by different methods using the same prompt and seed. The order of videos within each group is randomized. Participants are asked to select the best video based on overall visual quality and consistency, with the option to skip when samples are indistinguishable. As shown in Fig.~\ref{fig:user_study}, \methodname achieves the highest preference rate by a large margin, accounting for 53.5\% of total wins. In contrast, the second-best method Epipolar-DPO receives 22.4\% of the votes. These results indicate that the geometric improvements introduced by \methodname are consistently perceptible to humans and translate into higher preference.


\subsection{Qualitative Analysis}
We observe consistent qualitative improvements across several key aspects: i) \textit{Structural and geometric consistency} is substantially improved, with \methodname suppressing object splitting and preserving rigid-body integrity under camera motion (Fig.~\ref{fig:i2v teaser}a). ii) \textit{Texture stability} is enhanced, significantly reducing flickering in high-frequency regions such as building facades and fan grilles (Fig.~\ref{fig:i2v teaser}b). iii) \textit{Robustness under challenging lighting} is improved, preventing scene degradation in low-light conditions and stabilizing specular reflections on reflective surfaces (Fig.~\ref{fig:i2v teaser}c). iv) \textit{Object attribute consistency} is better maintained, with the model preserving semantic identity across frames, including color constancy and material appearance (Fig.~\ref{fig:i2v teaser}d). v) Although our approach does not explicitly optimize motion dynamics, it avoids degradation and often improves \textit{dynamic motion coherence}, maintaining object integrity in scenes involving large camera motion or dynamic elements (Appendix~\ref{sec:dyna}), which we further discuss in Sec.~\ref{subsec:manifold_hypothesis}. A comprehensive set of additional visualizations and frame-by-frame comparisons is provided in Appendix~\ref{sec:qualititive_results}.








%% file: sections/50-discussion.tex
\section{Discussion}
\label{sec:discussion}
\subsection{Scene-Level Geometry \textit{v.s.} Local Constraints}
In this subsection, we analyze why scene-level geometric preference modeling, as used in \methodname, provides a more reliable alignment signal than local, pairwise geometric constraints. Epipolar-DPO~\cite{kupyn2025epipolar} relies on frame-level epipolar relations, which are effective for minor geometric corrections but fragile under severe generative artifacts. In practice, local metrics can yield false positives, since degenerate outputs such as texture collapse or frozen regions may still satisfy sparse epipolar constraints or exhibit high local similarity, resulting in weak or misleading preference signals during alignment.

\begin{figure}[t]
  \centering
  \includegraphics[width=\linewidth ]{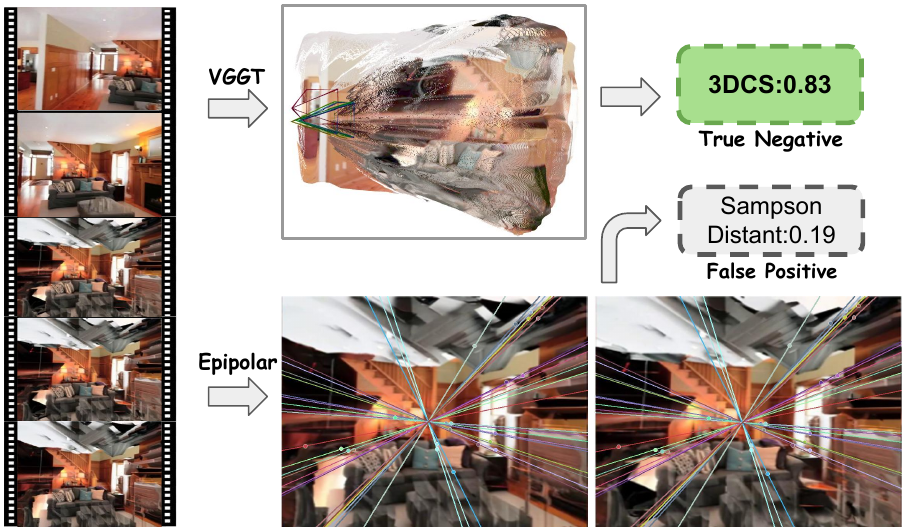}
   \caption{\textbf{Scene-level \textit{v.s.} local geometry.} Comparison between local geometric metric and scene-level metric on a corrupted video. Local, pairwise constraints yield a false positive, while the scene-level metric correctly identifies geometric inconsistency.} 
   \vspace{-5pt}
  \label{fig:false_positive}
\end{figure}

In contrast, \methodname evaluates geometric consistency at the scene-level by enforcing a global reprojection constraint. All frames must jointly admit a single, coherent 3D explanation. Otherwise, reconstruction error increases sharply. This global requirement prevents spatial drift from accumulating over time and provides a dense, unambiguous signal that penalizes collapsed or physically implausible generations. Consequently, preference optimization guided by scene-level geometry avoids rewarding locally consistent but globally invalid samples, leading to more stable alignment. Fig.~\ref{fig:false_positive} illustrates representative cases where epipolar-based metrics fail to penalize corrupted sequences that are correctly rejected by the proposed scene-level metric.

\subsection{Geometry as a Regularizer for Motion Generation}
\label{subsec:manifold_hypothesis}
\begin{figure}[t]
  \centering
  \includegraphics[width=\linewidth]{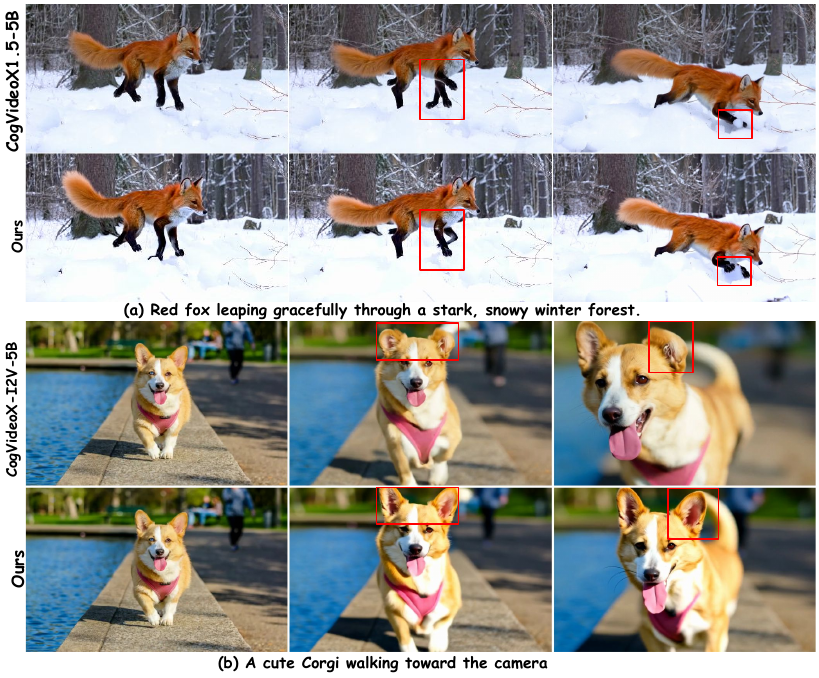}
   \caption{\textbf{Improved object motion coherence} in both T2V (top) and I2V (bottom) generation. VideoGPA better preserves object structure and motion continuity across frames.
} \vspace{-5pt}
  \label{fig:dyna_teaser}
\end{figure}

Although VideoGPA explicitly targets geometric consistency in predominantly static scenes, we observe consistent improvements in dynamic motion coherence as demonstrated in Fig.~\ref{fig:dyna_teaser} and reflected by higher Motion Quality (MQ) win rates in Table~\ref{tab:final_3d_comparison}. We interpret this behavior through the lens of the \textit{video motion manifold}. Video diffusion models strive to approximate the high-dimensional manifold of natural video distributions~\cite{ho2020denoisingdiffusionprobabilisticmodels,song2021scorebasedgenerativemodelingstochastic}. Geometric inconsistencies (e.g., warping backgrounds or inconsistent perspectives) represent a divergence from the physically plausible subspace of this manifold~\cite{karras2022elucidatingdesignspacediffusionbased, blattmann2023alignlatentshighresolutionvideo}.

 By enforcing strict geometric constraints, \methodname effectively acts as a \textit{geometric regularizer}, projecting the generative process back onto a manifold subspace where 3D consistency holds. This geometric stability serves as a foundational \textit{anchor} for dynamic generation. When the consistent background and camera trajectory adhere to projective geometry, the model can more effectively disentangle camera movement from object motion~\cite{wang2024motionctrlunifiedflexiblemotion,guo2024animatediffanimatepersonalizedtexttoimage}. Consequently, the model's capacity is freed from hallucinating spatial corrections, allowing its inherent motion priors to focus on generating coherent and realistic object dynamics. In essence, by fixing the geometry of the \textit{stage}, we enable the model to generate more coherent performances for the \textit{actors}. Futher discussion in Appendix~\ref{sec:dyna}.

%% file: sections/70-conclusion.tex
\section{Conclusion}


We address geometric inconsistency in video diffusion models through post-training alignment guided by a scene-level 3D consistency score derived from geometry foundation models. 
By integrating this score into a preference optimization framework, we demonstrate that pretrained video diffusion models can be effectively aligned to produce videos with substantially improved 3D coherence and temporal stability, without introducing explicit structural priors or degrading general-purpose generation quality. Our results suggest that geometric failures in current video generators are largely attributable to objective misalignment rather than architectural limitations, and can be mitigated through lightweight post-training. 
A remaining limitation is the scalability of geometric reconstruction, whose runtime and memory costs grow with video length. We expect advances in lightweight geometric foundation models to address this.

\section*{Acknowledgement}
The USC Physical Superintelligence Lab acknowledges generous support from Toyota Research Institute, Dolby, Google DeepMind, Capital One, Nvidia, and Qualcomm. Junjie Ye is supported by a fellowship from Capital One. Yue Wang is also supported by a Powell Research Award.

\section*{Impact Statement}
This paper presents work whose goal is to advance the field of machine learning. There are many potential societal consequences of our work, none of which we feel must be specifically highlighted here.

\newpage

%% file: sections/appendix.tex
\newpage
\appendix
\onecolumn

\section{Additional Details on Preference Data Construction}
\label{sec:appendix_data}

This appendix provides additional implementation details for the data curation pipeline used in post-training alignment. The core strategy is shared across both image-to-video (I2V) and text-to-video (T2V) settings, with additional design choices introduced for I2V to elicit diverse camera motion.

For both I2V and T2V alignment, we generate candidate videos by sampling from pretrained video diffusion models using multiple random seeds per conditioning input. Specifically, we source approximately 3,000 conditioning inputs from subsets $8k$, $9k$, $10k$, and $11k$ of the DL3DV-10K~\cite{ling2023dl3dv10klargescalescenedataset} dataset. For each input, we generate three video samples with 3 different random seeds, resulting in roughly 9,000 candidate videos in total. Detailed statistics for each model variant and training configuration are reported in Table~\ref{tab:training_details}.

Candidate samples are ranked using the 3D consistency score described in Sec.~\ref{sec:consistency score}, and preference pairs $(x_w, x_l)$ are constructed by selecting the best and worst samples within each group. To ensure the quality and discriminative power of the preference signal, we apply the same multi-stage filtering procedure for both I2V and T2V data.
\begin{itemize}
    \item \textbf{Motion Salience Filtering}: To avoid trivial static generations, we first quantify camera motion magnitude from reconstructed poses. For consecutive camera poses $(R_i, t_i)$, we compute translation $\Delta t_i = \|t_{i+1}-t_i\|_2$ and rotation $\Delta\theta_i=\arccos\!\left(\frac{\mathrm{Tr}(R_{i+1}R_i^{\top})-1}{2}\right)$. Translation is normalized by its mean step
    $s_{\mathrm{trans}}=\frac{1}{T-1}\sum_i\Delta t_i$,
    while rotation remains unscaled:
    \begin{equation}
    \bar{t}=\tfrac{1}{T-1}\sum_i\frac{\Delta t_i}{s_{\mathrm{trans}}},
    \qquad
    \bar{r}=\tfrac{1}{T-1}\sum_i\Delta\theta_i.
    \label{eq:motion_avg}
    \end{equation}

The motion score $\alpha$ is defined as:
    \begin{equation}
    \alpha = \bar{t} + \lambda \cdot \bar{r} + \epsilon,
    \label{eq:motion_score}
    \end{equation}
    where $\lambda = 0.1$. Samples with $\alpha < 0.001$ are discarded.
    
    \item \textbf{Geometric Margin Selection}: Within each generation group, we retain only preference pairs whose 3D consistency scores differ by more than 0.05, ensuring that the preference signal reflects a meaningful geometric distinction.

    \item \textbf{Difficulty Pruning}: Finally, we remove pairs in which the preferred sample still exhibits poor global geometry, defined as a consistency score greater than 0.8. This prevents the model from learning from low-quality geometric references.
\end{itemize}
After filtering, we obtain a compact set of preference pairs for each model variant, with statistics reported in Table~\ref{tab:training_details}. For fair comparison, we adopt the same preference data construction pipeline for Epipolar-DPO~\cite{kupyn2025epipolar}, including identical candidate generation and filtering steps. The two methods differ in the geometric signal used to evaluate candidate consistency, with Epipolar-DPO relying on local, pairwise epipolar constraints rather than the proposed scene-level 3D consistency formulation.

\paragraph{Scripted Prompting for I2V.} In the I2V setting, we additionally introduce scripted camera-motion prompts to elicit diverse camera trajectories while keeping scene content fixed. Each prompt consists of a static-scene constraint followed by a multi-stage camera motion description composed of $N \in \{2,3\}$ motion primitives. Motion primitives are drawn from three categories, translations, rotations, and complex paths, as listed in Table~\ref{tab:motion_primitives}, and concatenated using natural temporal connectors such as ``then'' and ``followed by''. An example of a generated prompt is: \textit{``slide sideways across the room, then pan toward the main subject, followed by a gentle roll to one side.''} The scripted camera-motion data construction procedure used to generate preference pairs is summarized in Algorithm~\ref{alg:data_prep}.

\begin{table}[!t]
\centering
\caption{The predefined motion primitives used for synthetic prompt generation. These actions are combined using temporal connectors to form diverse camera trajectories.}
\label{tab:motion_primitives}
\begin{small}
\begin{tabular}{lll}
\toprule
\textbf{Translations} & \textbf{Rotations} & \textbf{Complex Paths} \\
\midrule
push forward into the scene & pan across the room & orbit around the scene \\
pull back away from the scene & pan toward the main subject & arc around the center of the room \\
slide sideways across the room & scan across the shelves & circle around the main object \\
move laterally along the furniture line & tilt upward toward the ceiling & swing around the room \\
drift across the space & tilt downward toward the floor & pivot around the viewpoint \\
glide toward the room center & roll gently to one side & \\
shift through the foreground & look around the environment & \\
move diagonally through the space & & \\
\bottomrule
\end{tabular}
\end{small}
\end{table}

To prevent the model from introducing unintended dynamics, we prepend a fixed static-scene constraint to every prompt:
\begin{quote}
\textit{``A realistic continuation of the reference scene. Everything must remain completely static: no moving people, no shifting objects, and no dynamic elements. Only the camera is allowed to move.''}
\end{quote}
This constraint ensures that geometric consistency is the primary factor distinguishing candidate samples during alignment.

Please note that although this scripted setup explicitly restricts training samples to static scenes, we observe that the resulting alignment improves 3D consistency even in dynamic scenes generated by the model. As discussed in Sec.\ref{subsec:manifold_hypothesis}, this behavior suggests that enforcing scene-level geometric consistency provides a stabilizing inductive bias that transfers beyond the static training setting.

\begin{algorithm}[!t]
\caption{Scripted Camera-Motion Preference Data Construction for I2V}
\label{alg:data_prep}
\begin{algorithmic}[1]
\REQUIRE Set of first frames $\mathcal{I}$, motion primitive sets $\mathcal{M}_{T, R, C}$
\ENSURE Metadata repository $\mathcal{D}$
\FOR{each image $I_i \in \mathcal{I}$}
    \STATE $n \leftarrow \text{random}(\{2, 3\})$
    \STATE $\text{segments} \leftarrow \text{sample } n \text{ pieces from } \{\mathcal{M}_T \cup \mathcal{M}_R \cup \mathcal{M}_C\}$
    \STATE $\mathcal{A}_{motion} \leftarrow \text{JoinSegments}(\text{segments}, \text{``then''}, \text{``followed by''})$
    \STATE $P_{text} \leftarrow \text{StaticPrefix} + \text{``Camera motion: ''} + \mathcal{A}_{motion}$
    \STATE \COMMENT{Prune pairs with static motion, low-quality winners, or negligible gaps.}
    \STATE $\mathcal{D}[ID_i] \leftarrow \{I_i, \mathcal{A}_{motion}, P_{text}\}$
\ENDFOR
\STATE \textbf{return} $\mathcal{D}$
\end{algorithmic}
\end{algorithm}

\section{Training Details and GPU Usage}
\label{sec:GPU}
All experiments follow a unified optimization protocol to ensure fair comparison across methods. The training is conducted on $8 \times$ NVIDIA A100 GPUs using the AdamW optimizer with a peak learning rate of $5 \times 10^{-6}$. We employed a cosine decay schedule with 500 warm-up steps and a global batch size of 16. All post-training alignment uses LoRA~\cite{hu2021loralowrankadaptationlarge} with rank $r=64$ and $\alpha=128$, affecting approximately 1\% of the total model parameters.

\paragraph{Data Statistics} For preference-based alignment, we generate multiple candidate videos per conditioning input and curate preference pairs using the filtering strategy described in Sec.~\ref{subsec:preference_data} and Appendix~\ref{sec:appendix_data}. After filtering for motion salience and geometric margin, each model variant is trained on a compact set of $\sim 2500$ preference pairs. The same data construction pipeline is applied to \methodname\ and Epipolar-DPO to ensure fair comparison.

For the SFT baseline, we fine-tune on video-caption pairs derived from DL3DV-10K using captions generated by CogVLM2-Video, resulting in 20,356 training videos. This setup mirrors standard supervised fine-tuning without introducing additional geometric signals. A summary of dataset statistics, training steps, and wall-clock time is provided in Table~\ref{tab:training_details}.


\begin{table}[ht]
\centering
\caption{Statistics of curated datasets and training computational costs on 8$\times$A100 GPUs.}
\label{tab:training_details}
\begin{small}
\begin{tabular}{lcccc}
\toprule
\textbf{Model Variant} & \textbf{Initial Samples} & \textbf{Pref. Pairs} & \textbf{Steps} & \textbf{Total Time} \\
\midrule
CogVideoX-I2V-5B (\methodname) & 9,441 & 2,663 & 10,000 & \multirow{4}{*}{5d} \\
CogVideoX-I2V-5B (Epipolar-DPO) & 9,441 & 2,542 & 10,000 & \\
CogVideoX-T2V-5B (\methodname) & 8,496 & 2,550 & 10,000 & \\
CogVideoX-T2V-5B (Epipolar-DPO) & 8,496 & 2,330 & 10,000 & \\
\midrule
CogVideoX1.5-T2V-5B (\methodname) & 9,441 & 3,051 & 1,500\textsuperscript{*} & 3d \\
\midrule
SFT (T2V) & \multirow{2}{*}{20,356 (Clips)} & - & 10,000 & \multirow{2}{*}{2d}\\
SFT (I2V) & & - & 10,000 &  \\
\bottomrule
\end{tabular}
\end{small}

\vspace{2pt}
\raggedright \footnotesize \textsuperscript{*}For CogVideoX1.5-5B, we utilized a shortened schedule consisting of 500 warm-up steps followed by 1,000 optimization steps.
\end{table}

\section{Scalability and VRAM Consumption}\label{exp:scale_and_vram}
We analyze the scalability of the proposed 3D consistency score by varying the number of frames $T$ used for reconstruction. As summarized in Table~\ref{tab:mn3d_scaling_efficiency}, both runtime and memory consumption increase monotonically with sequence length. While throughput remains competitive for short clips, longer sequences incur higher computational cost and substantially increased VRAM usage.

In our experiments, we empirically select $T=10$ frames for post-training alignment. This choice is motivated by the temporal length of the base video models, which generate 49 frames (CogVideoX) or 81 frames (CogVideoX1.5). Sampling 10 frames already provides sufficiently dense temporal coverage to capture scene-level geometry, and we do not observe significant gains in geometric consistency when increasing $T$ beyond this range, while incurring substantially higher computational and memory costs.

More broadly, post-training alignment for long video generation remains challenging, as larger temporal windows require proportionally higher memory and compute. Addressing this limitation will likely require advances in more efficient geometric foundation models or scalable reconstruction strategies, and we leave this direction for future work.


\begin{table}[h]
\caption{
3D Consistency Score (VGGT): Performance and VRAM scalability analysis by frame count.}

\label{tab:mn3d_scaling_efficiency}
\begin{center}
\begin{tabular}{cccc}
\toprule
Frames & Avg Time (s) $\downarrow$ & FPS $\uparrow$ & Peak VRAM (GB) \\
\midrule
5 & 0.3785 & 13.21 & 11.63 \\
10 & 0.8695 & 11.50 & 13.86 \\
20 & 2.1460 & 9.32 & 19.84 \\
40 & 5.9529 & 6.72 & 32.58 \\
\bottomrule
\end{tabular}
\end{center}
\end{table} 

\paragraph{Latency and Throughput Comparison:}
We assessed the runtime efficiency of 3D consistency score against standard epipolar-based geometric metrics, including Epipolar Sampson Distance using \textsc{Sift} (CPU-based) and \textsc{LightGlue} (GPU-based). All experiments are conducted on a single NVIDIA RTX 6000 Ada GPU using sequences of $T=10$ frames. 
As shown in Table~\ref{tab:efficiency_baseline}, our dense consistency metric achieves comparable throughput to GPU-based epipolar methods (11.50 FPS vs.\ 12.10 FPS), despite operating on dense reconstructions rather than sparse correspondences. This demonstrates that reconstruction-based geometric signals can be computed efficiently and are practical for large-scale preference construction.

\input{tables/runtime_efficiency}

\section{Analysis of Scene and Motion Fidelity}
\label{sec:scene_complexity}
 A common concern in preference fine-tuning is that the model might "collapse" toward simpler, more predictable generations to satisfy specific reward constraints, leading to a loss of texture or movement. To mitigate this, VideoGPA incorporates \textbf{KL regularization} within the Direct Preference Optimization (DPO) framework, preventing the model from deviating excessively from the original policy's distribution.

To provide rigorous quantitative evidence that geometric alignment does not come at the cost of visual richness, we evaluate four spatio-temporal metrics across both Image-to-Video (I2V) and Text-to-Video (T2V) settings:
\begin{itemize}
    \item \textbf{Laplacian Variance (LV):} Evaluates texture sharpness and image clarity.
    \item \textbf{FFT High-Frequency Ratio (FHR):} Captures the preservation of fine-grained spatial details.
    \item \textbf{Edge Density (ED):} Indicates structural and compositional complexity.
    \item \textbf{Optical Flow Magnitude (OFM):} Quantifies motion intensity and dynamic range.
\end{itemize}

As demonstrated in Table~\ref{tab:complexity_metrics}, all four metrics are either strictly preserved or significantly improved after VideoGPA fine-tuning. Notably, in the I2V setting, we observe a substantial increase in Laplacian Variance (from $299.56$ to $839.26$), suggesting that our alignment process not only maintains but actually enhances the visual crispness of the generated frames. These results confirm that VideoGPA successfully achieves geometric consistency without suffering from scene complexity or motion degeneration.
\input{tables/scene_complexity}

\section{Ablation Study on Training Steps}
We analyze the effect of post-training duration in the I2V setting by evaluating models fine-tuned for different numbers of optimization steps. For each checkpoint, we prompt the model to generate 150 video samples and compute the metrics reported in Table~\ref{tab:ablation_steps_no_filter}. To ensure consistency with the main experiments, results from the final 10{,}000-step model are used for comparison with other methods.

As shown in Table~\ref{tab:ablation_steps_no_filter}, the model achieves competitive geometric and perceptual performance as early as 1{,}000 training steps, with only marginal improvements observed at later checkpoints. Notably, evaluation is conducted using natural, descriptive prompts that differ from the scripted camera-motion prompts used during I2V training, making the validation setting more challenging. Therefore, the limited gains at later stages indicate early convergence of the alignment objective under distribution shift, rather than overfitting to the training prompt format.


\begin{table}[h]
    \centering
    \caption{Ablation study on the number of post-training steps in I2V setting.}
    \label{tab:ablation_steps_no_filter}
    \setlength{\tabcolsep}{3pt}
    \begin{tabular}{lcccccc}
        \toprule
        \multirow{2}{*}{\textbf{Step}} & \multicolumn{3}{c}{\textbf{3D Reconstruction Error}} & \multicolumn{3}{c}{\textbf{3D Consistency}} \\
        \cmidrule(lr){2-4} \cmidrule(lr){5-7}
         & \textbf{PSNR} $\uparrow$ & \textbf{SSIM} $\uparrow$ & \textbf{LPIPS} $\downarrow$ & \textbf{MVCS} $\uparrow$ & \textbf{3DCS} $\downarrow$ & \textbf{Epipolar} $\downarrow$ \\
        \midrule
        Baseline & 18.4336  & 0.6176 & 0.5371  & 0.9662 &  0.5587 & 0.5819 \\
        Step-1000  & 19.5250 & 0.6648 & 0.5161 & \textbf{0.9829} & 0.5313 & 0.5264 \\
        Step-5000  & 19.5244 & 0.6647 & 0.5160 & \textbf{0.9829} & 0.5312 & 0.5264 \\
        Step-8000  & 19.5250 & 0.6648 & 0.5161 & \textbf{0.9829} & 0.5313 & 0.5264 \\
        Step-10000 & \textbf{19.6853} & \textbf{0.6756} & \textbf{0.5123} & 0.9822 & \textbf{0.5270} & \textbf{0.5229} \\
        \bottomrule
        \multicolumn{7}{l}{\footnotesize Note: Results are evaluated with VGGT backbone.} 
    \end{tabular}
\end{table}

\subsection{Limitations of Supervised Fine-Tuning for Geometric Grounding}

A common paradigm in preference learning is to initialize the policy with Supervised Fine-Tuning (SFT). However, our empirical evidence in Table~\ref{tab:sft_dpo_ablation} challenges this assumption for the task of geometric grounding. We compare VideoGPA (DPO-only) against two SFT baselines: (1) SFT on preference winners and (2) a large-scale SFT on 10,000 real-world videos followed by 1,000 steps of DPO.

Our results yield a striking observation: \textbf{SFT stages consistently harm the acquisition of geometric priors.} Specifically:
\begin{itemize}
    \item \textbf{DPO vs. SFT on Identical Data:} When trained on the same 2,500 winner samples, SFT-1K performs significantly worse than VideoGPA, with Epipolar error increasing from $0.509$ to $0.589$. This confirms that imitating "correct" samples is insufficient; the model requires the contrastive penalty provided by DPO to actively distinguish physically plausible motion from geometric artifacts.
    
    \item \textbf{Sequential Interference:} Even with an extensive SFT stage on 10,000 high-quality real-world videos (SFT$_{10K} \rightarrow$ DPO$_{1K}$), the performance fails to match the DPO-only configuration. This suggests that SFT on real-world data provides redundant signals that the foundation model has already internalized during pre-training. More critically, the SFT stage appears to \textit{prematurely constrain} the policy, creating an optimization barrier that hinders the subsequent DPO phase from discovering optimal 3D-consistent representations.
\end{itemize}

These findings indicate that for complex structural constraints like 3D geometry, \textbf{direct preference optimization is superior to conventional sequential pipelines.} The imitative nature of SFT lacks the discriminative power necessary for stable geometric grounding, making it less effective—and even counter-productive—compared to the direct distillation of geometric priors via VideoGPA.

\input{tables/sft_dpo}

\subsection{Foundation Model Backbone Robustness }

We further investigate whether the choice of the geometric foundation model backbone influences the preference signals. Specifically, we evaluate the consensus between \textit{VGGT} and alternative backbones, \textit{DUSt3R}~\cite{dust3r} and \textit{Depth Anything V3 (DA3)}~\cite{lin2025depth3recoveringvisual}, across over 3,000 I2V groups. To quantify this, we define three hierarchical metrics:
\begin{itemize}
    \item \textbf{Agreement Rate (AR):} The percentage of pairs where the backbone agrees with VGGT on the binary winner/loser relationship.
    \item \textbf{Top-1 Consistency (T1):} The probability that the best-performing sample identified by VGGT is also ranked first by the alternative backbone within a triplet.
    \item \textbf{Full Ranking Agreement (FR):} The rate at which the entire ordinal ranking of a group remains identical across backbones.
\end{itemize}

As shown in Table~\ref{tab:backbone_agreement}, DA3 achieves 100\% consensus with VGGT across all 3,000+ groups. This high degree of alignment suggests that as geometric foundation models evolve, their preference signals converge toward a consistent 3D geometric interpretation, validating the reliability of our supervision source.

\input{tables/backbone}

Furthermore, to test the robustness of our DPO-based training against potential supervision noise, we conducted a stress test by randomly flipping up to $20\%$ of the preference labels for 1,000 steps. As reported in Table~\ref{tab:noise_tolerance}, the model maintains strong performance even with a $20\%$ noise injection, with only a marginal decline in overall preference (OVL). This demonstrates that our pipeline is highly resilient to occasional labeling errors, ensuring stable geometric alignment in practical scenarios.

\input{tables/tolerance}

\section{Generalization on Out-of-Distribution Datasets}
\label{sec:ood_evaluation}

To evaluate the zero-shot generalization capability of VideoGPA, we conduct experiments on two distinct out-of-distribution (OOD) datasets: \textit{WebVid}~\cite{bain2021frozen} and \textit{Panda-70M}~\cite{chen2024panda}. These datasets are disjoint from our training set and feature significantly different visual distributions and complex motion dynamics.

\subsection{Quantitative Performance on WebVid}
We evaluate 100 randomly sampled videos from WebVid~\cite{bain2021frozen}, comparing VideoGPA against the baseline, SFT, and Epipolar-DPO. As shown in Table~\ref{tab:webvid}, VideoGPA achieves the best performance in \textit{LPIPS}, \textit{3DCS}, and \textit{Epipolar} error. These results confirm that our geometric consistency improvements effectively generalize to OOD web-scale content.

\input{tables/ood_quatitative_evaluation}

\subsection{Generalization to Motion Coherence (Panda-70M)}
A critical challenge in video generation is maintaining \textbf{motion coherence} in highly dynamic scenes. We tested VideoGPA on 100 dynamic videos from \textit{Panda-70M}~\cite{chen2024panda}, which contain complex object movements and camera transitions absent from our relatively static-scene training data. 

As summarized in Table~\ref{tab:panda_videoreward}, VideoGPA significantly outperforms the baseline across all VideoReward metrics, including a substantial gain in \textit{Overall Preference} (OVL: 64.0\%). This improvement on OOD dynamic sequences suggests that VideoGPA does not merely memorize static geometric patterns; instead, it enforces a \textbf{temporally coherent geometric prior} that stabilizes motion even in highly dynamic contexts. We will further explore in Appendix~\ref{sec:dyna}.

\input{tables/pandas}

\section{Additional Evaluation on Wan Family}
\label{sec:wan}

To examine whether \methodname generalizes to other video diffusion architectures, we evaluate it on \textbf{Wan2.2-TI2V-5B}~\cite{wan2025wanopenadvancedlargescale}. We follow the same protocol used for our CogVideoX experiments: preference pairs are constructed from videos generated under scripted static-scene camera-motion prompts (Sec.~\ref{sec:appendix_data}).

\paragraph{Static-scene camera-motion evaluation.} The picture changes once we 
evaluate Wan2.2 on the same scripted static-scene prompts used during preference 
data construction, a setting that explicitly stresses 3D geometric consistency 
under camera motion. As shown in the top half of Table~\ref{tab:wan}, the base 
Wan2.2 model degrades noticeably in this regime, and \methodname recovers 
substantial performance: PSNR improves from $19.40$ to $24.41$, LPIPS drops 
from $0.480$ to $0.394$, and Epipolar error reduces from $0.594$ to $0.499$, 
alongside consistent gains across all 3D consistency metrics. These geometric improvements also translate to stronger human-aligned quality, 
with \methodname achieving VideoReward win rates of $52.00\%$ in Visual Quality (VQ), 
$65.00\%$ in Motion Quality (MQ), and $57.00\%$ in Overall preference (OVL).

\paragraph{VLM generated caption evaluation.}
As shown in the bottom half of Table~\ref{tab:wan}, under natural generated video 
captions where the base model is already strong, VideoGPA still yields modest 
improvements in reconstruction quality and most consistency metrics, including 
PSNR ($24.06\rightarrow24.41$), SSIM ($0.815\rightarrow0.823$), LPIPS 
($0.412\rightarrow0.401$), MVCS ($0.941\rightarrow0.945$), and 3DCS 
($0.424\rightarrow0.411$), though Epipolar error shows a slight increase 
($0.588\rightarrow0.611$). In terms of human-aligned quality, \methodname 
achieves VideoReward win rates of $47.00\%$ in Visual Quality (VQ), $67.00\%$ 
in Motion Quality (MQ), and $56.00\%$ in Overall preference (OVL). Together, 
these results suggest that VideoGPA provides the most pronounced geometric 
benefits in the camera-motion regime where modern video models remain weakest, 
while consistently improving performance under unconstrained natural language 
prompts.

\input{tables/wan}

\section{Emergent Consistency in Dynamic Scenes}
\label{sec:dyna}

\begin{figure}[h]
  \centering
  \includegraphics[width=0.99\linewidth]{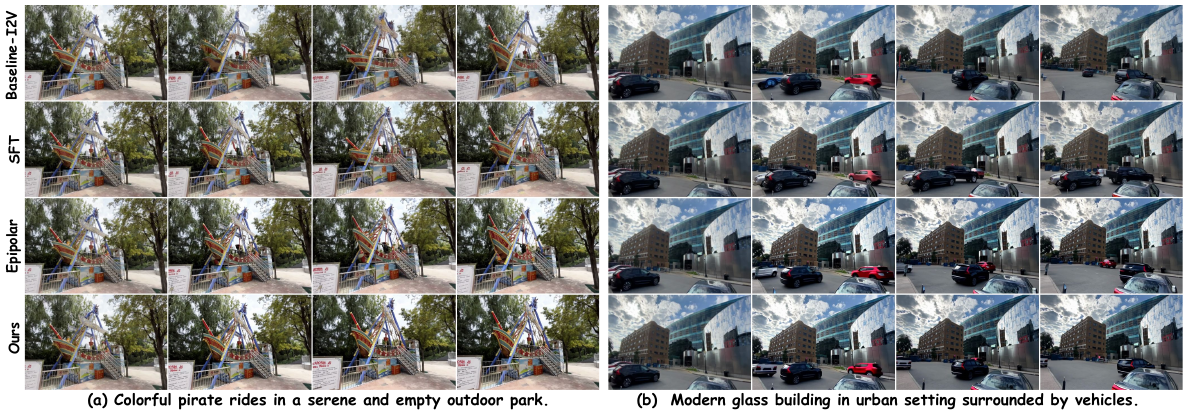}
   \caption{\textbf{Examples of dynamic object generation in unconstrained scenes.} When prompted with dynamic object without static scene constraints, (a) \methodname is the only approach that preserves the rigid-body integrity of the pirate ride during spinning. (b) Our method successfully maintains the physical plausibility of the vehicle throughout its trajectory and ensures color consistency for the red car after occlusion occurs, whereas baseline models exhibit color drift or structural distortion.} 
  \label{fig:naive_dyna}
\end{figure}
In our validation tests, we observe that when the static-scene constraint is relaxed, the model occasionally generates moving objects based on their natural semantics, such as a spinning pirate ride or moving vehicles. In these cases, \methodname significantly improves the coherence of object motion, even in complex scenarios that differ substantially from the static scenes used during training. As illustrated in Fig.~\ref{fig:naive_dyna}, when prompted with dynamic subjects, the model exhibits superior \textbf{temporal stability} and \textbf{physical plausibility}.

\textbf{The Disentanglement Hypothesis} Initially, we hypothesized that these improvements might arise from a trivial background-foreground disentanglement effect. We posited that by enforcing background rigidity, the model was essentially ``faking" coherent motion by constraining foreground objects to satisfy local multi-view consistency against a stabilized background.

\textbf{Experimental Discovery.} However, our extensive experiments in both Image-to-Video (I2V) and Text-to-Video (T2V) settings provided results that overturned this initial intuition. As demonstrated in Sec.~\ref{subsection:Dynamic Scene}, the observed improvements represent a motion-coherent level advancement rather than a simple static multi-view consistency fix. Remarkably, even in the presence of highly articulated or non-rigid motions, such as animal limb movements, ear morphing, \methodname effectively suppresses chronic artifacts like ``geometry collapse" and ``semantic metamorphosis." These findings provide compelling empirical evidence for our \textbf{Video Motion Manifold} hypothesis (Sec.~\ref{subsec:manifold_hypothesis}). By regularizing the model toward a 3D-consistent manifold, we inherently facilitate more physically-grounded generative behaviors. Even historically "hard" cases, such as intricate limb inconsistency or objects splitting and merging, are rectified. This proves that our geometric alignment effectively reshapes the underlying motion manifold of the video diffusion model.

\textbf{Further Discussion: Physical Plausibility Improvement.}

Additionally, we observe that \methodname significantly enhances the \textbf{physical plausibility} of the generated videos. We hypothesize that this improvement is a beneficial side effect of distilling knowledge from the 3D reconstruction model. Since VGGT~\cite{wang2025vggt} is pre-trained on large-scale 3D datasets and DINO~\cite{caron2021emergingpropertiesselfsupervisedvision,oquab2024dinov2learningrobustvisual} possesses strong visual understanding and segmentation capabilities, the reconstruction model exhibits a strong geometric prior. This bias encourages the generation of point clouds that align with physically plausible object structures. Consequently, leveraging this geometric signal as a ranking criterion guides the model toward producing objects with more realistic physical properties, a capability fundamentally absent in frame-level constraint such as Epipolar~\cite{kupyn2025epipolar} or Vision-Language-Model based metrics~\cite{li2025videohalluevaluatingmitigatingmultimodal,li2025surveystateartlarge}.

\subsection{Qualitative Analysis of Complex Motion Coherence}
The following qualitative comparisons evaluate motion coherence in animal movements across our I2V and T2V experimental settings. These scenarios are widely considered the most challenging benchmarks for both closed-source and open-source video generation models. Note that we exclude GeoVideo as its released checkpoint is specialized for static environments and does not yield plausible dynamic scene generation. Additional high-resolution results can be found on the anonymous project webpage, with the link provided as a separate file in the supplementary package.

\label{subsection:Dynamic Scene}
\foreach \i in {1,...,10}{
    \noindent 
    \begin{minipage}{\textwidth}
        \centering
        \includegraphics[width=0.99\linewidth]{visualization/dyna/page_\i.pdf}

    \end{minipage}

}

\section{Additional Qualitative Results}
\label{sec:qualititive_results}

\subsection{Image-to-Video}
We present additional image-to-video results in this subsection. Compared to the base model, SFT, and Epipolar-DPO, \methodname produces more stable geometry under camera motion, with reduced spatial drift and texture flickering.

\foreach \i in {1,2,...,15}{
    \noindent 
    \begin{minipage}{\textwidth}
        \centering
        \includegraphics[width=0.99\linewidth]{visualization/i2v/page_\i.pdf}

    \end{minipage}

}

\subsection{Text-to-Video on CogVideoX-5B}
We show additional text-to-video results in this subsection. \methodname better preserves object structure and appearance over time, reducing deformation and semantic drift across diverse, visually complex scenes.

\foreach \i in {1,...,10}{
    \noindent 
    \begin{minipage}{\textwidth}
        \centering
        \includegraphics[width=0.99\linewidth]{visualization/t2v/page_\i.pdf}
    \end{minipage}
}

%% file: tables/runtime_efficiency.tex
\begin{table}[h]
\caption{\textbf{Runtime efficiency comparison} between 3D consistency score and epipolar-based metrics on 10-frame sequences.}
\label{tab:efficiency_baseline}
\begin{center}
\begin{small}
\begin{sc}
\begin{tabular}{lcc}
\toprule
Method & Time (s/video) $\downarrow$ & FPS $\uparrow$ \\
\midrule
{\rm Epipolar} ({\textsc{Sift}}) & 3.91 & 2.56 \\
{\rm Epipolar} ({\textsc{LightGlue}}) & \textbf{0.83} & \textbf{12.10} \\
\midrule
{\rm \textbf{3D Consistency Score (VGGT)}} & 0.86 & 11.50 \\
\bottomrule
\end{tabular}
\end{sc}
\end{small}
\end{center}
\vspace{-8pt}
\end{table}

%% file: tables/scene_complexity.tex
\begin{table}[h]
\small
\centering
\caption{\textbf{Scene complexity analysis.} We report Laplacian Variance (LV), FFT High-frequency Ratio (FHR), Edge Density (ED), and Optical Flow Magnitude (OFM) to quantify scene detail and motion intensity. All metrics are preserved or improved after VideoGPA fine-tuning.}
\label{tab:complexity_metrics}
\begin{tabular}{lcccc}
\toprule
\textbf{Model} & \textbf{LV} $\uparrow$ & \textbf{FHR} $\uparrow$ & \textbf{ED} $\uparrow$ & \textbf{OFM} $\uparrow$ \\
\midrule
\multicolumn{5}{c}{Image-to-Video (I2V)} \\
\midrule
Baseline-I2V & 299.56 $\pm$ 217.91 & 0.506 $\pm$ 0.054 & 0.117 $\pm$ 0.053 & 10.25 $\pm$ 4.77 \\
\textbf{VideoGPA (Ours)} & \textbf{839.26 $\pm$ 687.60} & \textbf{0.558 $\pm$ 0.058} & \textbf{0.136 $\pm$ 0.059} & \textbf{11.17 $\pm$ 4.75} \\
\midrule
\multicolumn{5}{c}{Text-to-Video (T2V)} \\
\midrule
Baseline-T2V & 711.13 $\pm$ 374.30 & 0.585 $\pm$ 0.058 & 0.113 $\pm$ 0.049 & 11.46 $\pm$ 5.42 \\
\textbf{VideoGPA (Ours)} & \textbf{917.12 $\pm$ 512.53} & \textbf{0.594 $\pm$ 0.056} & \textbf{0.115 $\pm$ 0.049} & \textbf{11.55 $\pm$ 5.61} \\
\bottomrule
\end{tabular}
\end{table}

%% file: tables/sft_dpo.tex
\begin{table*}[!h]
\small
\centering
\caption{\textbf{Ablation on Optimization Strategies.} We evaluate whether SFT aids or hinders geometric alignment. (1) \textit{DPO vs. SFT}: On identical winner samples, DPO significantly outperforms SFT. (2) \textit{Sequential Interference}: Direct DPO training from the base model surpasses the SFT (10K steps on real videos) $\to$ DPO pipeline. These results suggest that \textbf{SFT stages may harm geometric priors} by prematurely constraining the policy, making it less effective than direct preference optimization for structural consistency.}
\label{tab:sft_dpo_ablation}
\begin{tabular}{lccccccc}
\toprule
\textbf{Config} & \textbf{PSNR} $\uparrow$ & \textbf{SSIM} $\uparrow$ & \textbf{LPIPS} $\downarrow$ & \textbf{MVCS} $\uparrow$ & \textbf{3DCS} $\downarrow$ & \textbf{Epipolar} $\downarrow$ & \textbf{OVL} $\uparrow$ \\
\midrule
CogVideoX-I2V-5B & 22.854 & 0.786 & 0.476 & 0.945 & 0.485 & 0.585 & -- \\
\midrule
SFT-1K (Winners) & 21.285 & 0.772 & 0.489 & \textbf{0.957} & 0.498 & 0.589 & 39.0\% \\
SFT (10K) $\to$ DPO (1K) & \textbf{23.115} & 0.811 & 0.458 & 0.951 & 0.464 & 0.561 & 63.0\% \\
\textbf{DPO-1K} & 23.045 & \textbf{0.831} & \textbf{0.438} & 0.954 & \textbf{0.445} & \textbf{0.509} & \textbf{66.0\%} \\
\bottomrule
\multicolumn{4}{l}{\footnotesize {Reprojection based metrics are calculated with DA3-Large backbone.}} 
\end{tabular}
\end{table*}

%% file: tables/backbone.tex
\begin{table}[!h]
\small
\centering
\caption{\textbf{Preference Agreement Analysis.} We evaluate the consistency of alternative backbones against VGGT’s preference pairs across 3,000+ samples. DA3 demonstrates 100\% agreement.}
\label{tab:backbone_agreement}
\begin{tabular}{lccc}
\toprule
\textbf{Backbone} & \textbf{AR}  & \textbf{T1} & \textbf{FR} \\
\midrule
DUSt3R-ViTLarge & 80.8\% & 62.5\% & 48.5\% \\
DA3-Large & 100.0\% & 100.0\% & 100.0\% \\
\bottomrule
\end{tabular}
\end{table}

%% file: tables/tolerance.tex
\begin{table*}[h]
\small
\centering
\caption{\textbf{ Ablation on Geometric Signal Robustness. }Performance of VideoGPA under different levels of label noise. The consistent gains across all metrics, even with 20\% flipped labels, confirm the pipeline's tolerance to potential inaccuracies in geometric supervision.}
\label{tab:noise_tolerance}
\begin{tabular}{lccccccc}
\toprule
\textbf{Config} & \textbf{PSNR} $\uparrow$ & \textbf{SSIM} $\uparrow$ & \textbf{LPIPS} $\downarrow$ & \textbf{MVCS} $\uparrow$ & \textbf{3DCS} $\downarrow$ & \textbf{Epipolar} $\downarrow$ & \textbf{OVL} $\uparrow$ \\
\midrule
CogVideoX-I2V-5B & 22.854 & 0.786 & 0.476 & 0.945 & 0.485 & 0.585 & - \\
\midrule
00\% flipped & 23.045 & 0.831 & 0.438 & \textbf{0.954} & 0.445 & \textbf{0.509} & \textbf{66.0\%} \\
10\% flipped & \textbf{23.380} & \textbf{0.834} & \textbf{0.436} & 0.952 & \textbf{0.443} & 0.516 & \textbf{66.0\%} \\
20\% flipped & 23.253 & 0.832 & 0.439 & 0.951 & 0.446 & 0.523 & 63.0\% \\
\bottomrule
\multicolumn{4}{l}{\footnotesize {Reprojection based metrics are calculated with DA3-Large backbone.}} 
\end{tabular}
\end{table*}

%% file: tables/ood_quatitative_evaluation.tex
\begin{table*}[h]
\small
\centering
\caption{\textbf{Quantitative evaluation on WebVid.} We report 3D reconstruction and geometric consistency metrics. VideoGPA achieves the best LPIPS, 3DCS, and Epipolar scores, confirming that geometric consistency improvements generalize effectively across datasets.}
\label{tab:webvid}
\begin{tabular}{lcccccc}
\toprule
\textbf{Config} & \textbf{PSNR} $\uparrow$ & \textbf{SSIM} $\uparrow$ & \textbf{LPIPS} $\downarrow$ & \textbf{MVCS} $\uparrow$ & \textbf{3DCS} $\downarrow$ & \textbf{Epipolar} $\downarrow$ \\
\midrule
Baseline-I2V & 17.434 & 0.598 & 0.542 & 0.966 & 0.582 & 1.282 \\
SFT & \textbf{20.453} & 0.689 & 0.470 & \textbf{0.975} & 0.486 & 0.748 \\
Epipolar-DPO & 19.445 & \textbf{0.710} & 0.443 & 0.972 & 0.462 & 0.699 \\
\textbf{VideoGPA (Ours)} & 18.767 & 0.693 & \textbf{0.435} & 0.972 & \textbf{0.460} & \textbf{0.602} \\
\bottomrule
\multicolumn{7}{l}{\footnotesize {Reprojection based metrics are calculated with DA3-Large backbone.}} 
\end{tabular}
\end{table*}

%% file: tables/pandas.tex
\begin{table*}[!h]
\small
\centering
\caption{\textbf{VideoReward on Panda-70M.} Evaluated on 100 dynamic videos. VideoGPA significantly outperforms the baseline in zero-shot settings. This improvement in dynamic scenes (absent from static-scene training) supports the hypothesis that the model learns generalizable geometric priors.}
\label{tab:panda_videoreward}
\begin{tabular}{lcccc}
\toprule
\textbf{Metric} & \textbf{VQ} & \textbf{MQ} & \textbf{TA} & \textbf{OVL} \\
\midrule
Baseline-I2V & -- & -- & -- & -- \\
\textbf{VideoGPA (Ours)} & \textbf{61.0\%} & \textbf{61.0\%} & \textbf{51.0\%} & \textbf{64.0\%} \\
\bottomrule
\end{tabular}
\end{table*}

%% file: tables/wan.tex
\begin{table*}[h]
\small
\centering
\caption{Quantitative evaluation of \methodname on \textbf{Wan2.2-TI2V-5B} under two prompt regimes: (top)  scripted static-scene prompts with explicit camera-motion descriptions, which directly stress 3D geometric consistency; and (bottom) natural video captions generated by CogVLM2, on which the base model is already strong. \methodname substantially improves the base model under scripted camera-motion prompts and yields consistent, albeit modest, improvements under natural captions.}
\label{tab:wan}
\begin{tabular}{l|ccc|ccc|cccc}
\toprule
\multirow{2}{*}{\textbf{Method}} & \multicolumn{3}{c|}{\textbf{3D Reconstruction Error}} & \multicolumn{3}{c|}{\textbf{3D Consistency}} & \multicolumn{4}{c}{\textbf{VideoReward (Win Rate \%)}} \\
\cmidrule(lr){2-4} \cmidrule(lr){5-7} \cmidrule(lr){8-11}
 & \textbf{PSNR} $\uparrow$ & \textbf{SSIM} $\uparrow$ & \textbf{LPIPS} $\downarrow$ & \textbf{MVCS} $\uparrow$ & \textbf{3DCS} $\downarrow$ & \textbf{Epipolar} $\downarrow$ & \textbf{VQ} & \textbf{MQ} & \textbf{TA} & \textbf{OVL} \\

\midrule
\multicolumn{7}{l}{\textit{Scripted static-scene camera-motion prompts}} \\
\midrule
Baseline (Wan2.2) & 19.40 & 0.723 & 0.480 & 0.898 & 0.506 & 0.594 & - & - & - & -\\
\textbf{VideoGPA (Ours)} & \textbf{24.41} & \textbf{0.842} & \textbf{0.394} & \textbf{0.944} & \textbf{0.451} & \textbf{0.499} & \textbf{52.00} &\textbf{ 65.00 }& 35.00 & \textbf{57.00} \\
\midrule
\multicolumn{7}{l}{\textit{Natural CogVLM2-generated video captions}} \\
\midrule
Baseline (Wan2.2) & 24.06 & 0.815 & 0.412 & 0.941 & 0.424 & \textbf{0.588 } & - & - & - & -\\
\textbf{VideoGPA (Ours)} & \textbf{24.41} & \textbf{0.823} & \textbf{0.401} & \textbf{0.945} & \textbf{0.411} & 0.611 & 47.00 & \textbf{67.00} & 45.00 & \textbf{56.00} \\
\bottomrule
\multicolumn{7}{l}{\footnotesize {Reprojection based metrics are calculated with DA3-Large backbone.}}
\end{tabular}

\end{table*}